\documentclass[letterpaper, 10 pt, conference]{ieeeconf}
\IEEEoverridecommandlockouts
\let\labelindent\relax
\usepackage{enumitem}
\usepackage{balance}
\usepackage[font={small}]{caption}
\usepackage[labelformat=simple]{subfig}
\usepackage{array}
\usepackage{textcomp}
\usepackage{mathtools, nccmath}
\usepackage{graphicx}
\usepackage{amsfonts}
\usepackage{amsmath}
\usepackage{amssymb}
\usepackage{hyperref}
\usepackage{tikz}
\usepackage{arydshln}
\usepackage{multirow}
\usepackage{bm}
\usepackage{epstopdf}
\usepackage{cite}
\usepackage{siunitx}
\usepackage{xcolor}
\usepackage[ruled,linesnumbered,boxed,noend]{algorithm2e}
\usepackage{booktabs}
\usepackage{graphicx}
\usepackage{threeparttable}
\usepackage{stfloats}

\newcommand{\ob}[1]{\overline{\textbf{#1}}}
\newcommand{\tb}[1]{\tilde{\textbf{#1}}}

\title{\LARGE \bf PUTN: A Plane-fitting based Uneven Terrain Navigation Framework }
\author{ Zhuozhu Jian$^\dagger$, Zihong Lu$^\dagger$, Xiao Zhou, Bin Lan, Anxing Xiao, Xueqian Wang*, Bin Liang
\thanks{$^\dagger$ indicates equal contribution.}
\thanks{* Corresponding authors: Xueqian Wang.}
\thanks{This work is supported by Joint Funds of the National Natural Science Foundation of China U1813216.}
\thanks{Zhuozhu Jian, Bin Lan, Xueqian Wang, and Bin Liang are with the Center for Artificial Intelligence and Robotics, Shenzhen International Graduate School, Tsinghua University, Shenzhen 518055, China, \tt\{jzz21@mails., lan.bin@sz., wang.xq@sz., liangbin@\}tsinghua.edu.cn}
\thanks{Zihong Lu, and Xiao Zhou are with School of Mechanical Engineering and Automation at Harbin Institute of Technology, Shenzhen 518055, China. \tt \{200320802,180210129\}@stu.hit.edu.cn}
\thanks{Anxing Xiao is with Department of Electronic and Electrical Engineering at Southern University of Science and Technology, Shenzhen 518055, China, \tt xiaoax@mail.sustech.edu.cn}
}
\begin{document}
\maketitle
\begin{abstract}
 
Autonomous navigation of ground robots has been widely used in indoor structured 2D environments, but there are still many challenges in outdoor 3D unstructured environments, especially in rough, uneven terrains. This paper proposed a plane-fitting based uneven terrain navigation framework (PUTN) to solve this problem. The implementation of PUTN is divided into three steps. First, based on Rapidly-exploring Random Trees (RRT), an improved sample-based algorithm called Plane Fitting RRT* (PF-RRT*) is proposed to obtain a sparse trajectory. Each sampling point corresponds to a custom traversability index and a fitted plane on the point cloud. These planes are connected in series to form a traversable ``strip''. Second, Gaussian Process Regression is used to generate traversability of the dense trajectory interpolated from the sparse trajectory, and the sampling tree is used as the training set. Finally, local planning is performed using nonlinear model predictive control (NMPC). By adding the traversability index and uncertainty to the cost function, and adding obstacles generated by the real-time point cloud to the constraint function, a safe motion planning algorithm with smooth speed and strong robustness is available. Experiments in real scenarios are conducted to verify the effectiveness of the method. The source code is released for the reference of the community\footnote{Source code: \url{https://github.com/jianzhuozhuTHU/putn}.}.

\end{abstract}

\section{Introduction}
\label{sec:Introduction}
  
With the development of simultaneous localization and mapping (SLAM) technology and the improvement of computer performance, autonomous navigation technology is widely applied to ground robots. At present, the technology of 2D indoor ground mobile robots is relatively mature. By improving the robustness and efficiency of existing 2D autonomous navigation solutions \cite{putz2018move}\cite{bansal2020combining}\cite{kalogeiton2019real}, the autonomous navigation of robots has been widely used in many fields \cite{yuan2019multisensor}\cite{xiao2021robotic}, but autonomy on unstructured and off-road terrain remains a challenge. The difficulties mainly include the following two aspects: 1) the representation and calculation of large-scale maps are time-consuming; 2) unstructured terrain and real-time obstacles affect path planning.

To address these issues, this study first presents a new autonomous navigation framework based on point cloud plane fitting. It contains three parts: 1) sparse global trajectory generation based on random sampling combined with plane fitting; 2) dense path generation based on gaussian process regression (GPR) and linear interpolation; 3) a nonlinear model predictive control (NMPC) planner that avoids dynamic obstacle and guarantees safety. Our work focuses on the surface of the point cloud map, extracts a set of point clouds around the sampling node, and then fits a local plane with the set. For this plane, we propose custom evaluation indicators including flatness, slope, and sparsity, which are integrated to evaluate the traversability. Based on the fitted plane, we combine sampling, GPR, NMPC methods for motion planning of the robot.
   
   
\begin{figure}[t]
    \centering
    \includegraphics[width=8.5cm]{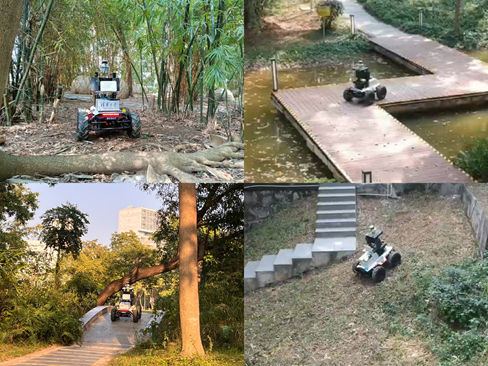}
    \caption{The Scout 2.0 four-wheel-drive platform is used to conduct experiments in steep slope, flat bridge, forest and arch bridge areas, respectively to verify the feasibility, robustness, and efficiency of PUTN.}
    \label{fig_exp_sen1}
    \vspace{-0.2cm}
\end{figure}

\subsection{Related Work}
\begin{figure*}[t]
    \centering
    \includegraphics[width=17.5cm]{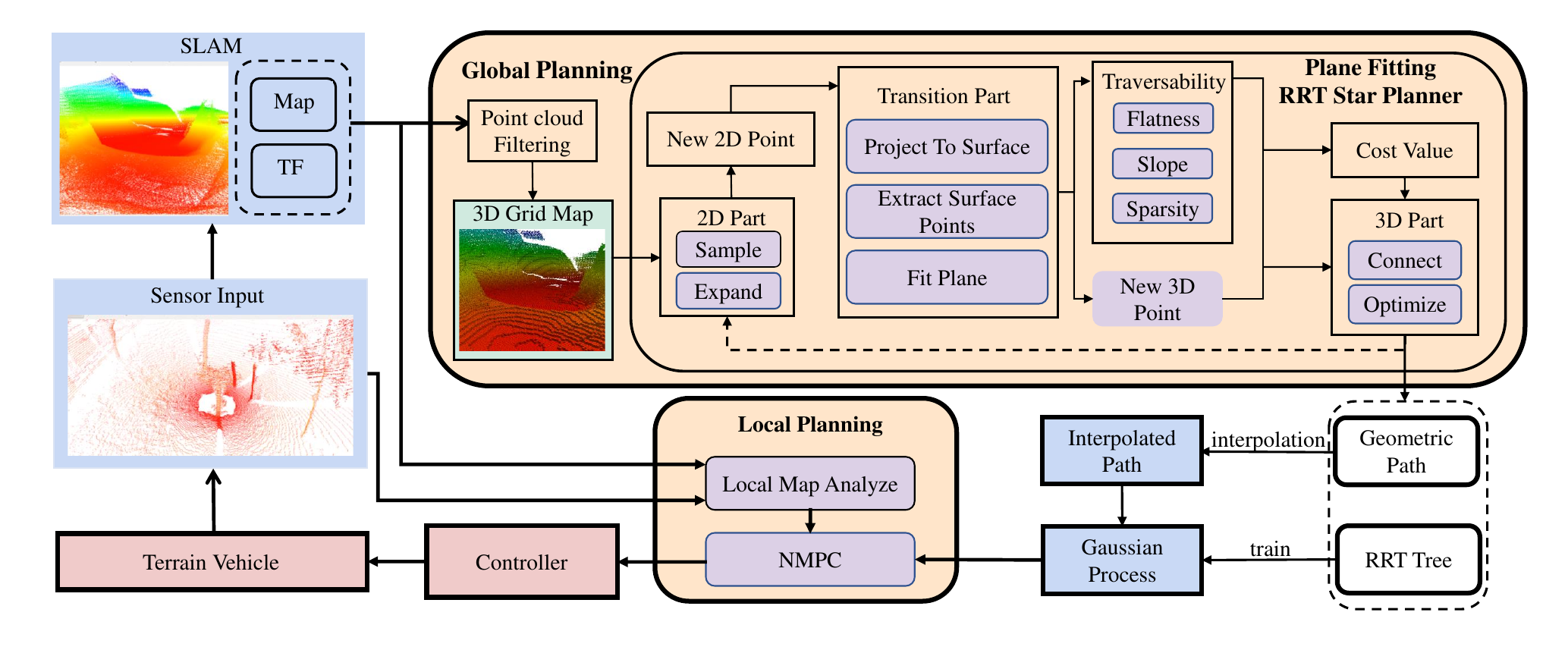}
    \caption{
    Overview of system framework of PUTN. From left to right: lidar detects surrounding terrain and returns point clouds. Then, the SLAM module builds a global map based on the point clouds. Next, the global map is imported to the Global Planning module, which uses the PF-RRT* algorithm to generate an RRT tree and a sparse global path. After interpolation based on GPR, a dense global path is imported to the Local Planning module based on NMPC Planner. The local planner produces control inputs to the robot.
    }
    \label{fig_framework_1}
    \vspace{-0.3cm}
\end{figure*}

Autonomous navigation based on unmanned ground vehicles (UGV) has been greatly developed in recent years. Based on 2D navigation, many frameworks for 3D navigation have been proposed. In \cite{wang2019safe}, a multi-layer 2D map extracted from the 3D OctoMap is used to explore staircases and slopes. However, this approach limits the ability to explore more complex environments. In \cite{kim2018slam} \cite{jeong2021motion} \cite{xiao2021learning}, global trajectory in more complex environmental scenarios is resolved, but their work all lack processing of real-time point clouds, which may lead to dangerous situations during the movement of the robot. Fan et al.\cite{fan2021step} apply rapid uncertainty-aware mapping and traversability evaluation to the robot, and tail risk is assessed using the Conditional Value-at-Risk (CVaR). The most visible deficiency is that when working in large-scale scenarios, the algorithm needs to analyze each point to generate a risk map, which consumes a large number of computing resources.

Map representation of space is essential for finding paths. The density of the map affects the accuracy and speed of planning \cite{papadakis2012constraint}. Therefore, when planning requires greater accuracy, the RGBD camera is more commonly used \cite{fankhauser2018probabilistic} \cite{campos2021autonomous} \cite{zhou2019robust}. In order to obtain a denser point cloud map with lidar, Shan et al.\cite{shan2018bayesian} apply Bayesian generalized kernel inference to terrain elevation and traversability inference. In this way, although theoretically higher accuracy can be obtained, it takes more time. In \cite{krusi2017driving}, the authors compute trajectories compliant with curvature and continuity constraints directly on unordered point cloud maps, omitting any kind of explicit surface reconstruction. But with this approach, the accuracy of the planning heavily depends on the density of the map.

\subsection{Contributions}
This work offers the following contributions: 
\begin{enumerate}
    \item Based on Rapidly-exploring Random Trees (RRT)\cite{lavalle2001randomized}, a new path planning algorithm named Plane Fitting RRT* (PF-RRT*) for uneven terrain integrating a new terrain assessment method is proposed.
    \item A new Plane-fitting based Uneven Terrain Navigation framework (PUTN) for navigation of ground robots on uneven terrain is proposed.
    \item Experiments in the real scenario are carried out to verify the real-time performance, effectiveness, and stability of the above algorithms.
\end{enumerate}

\section{Overview of the Framework}
\label{sec:framework}

\subsection{Problem Statement}
\label{subsec:problem}
\
Define $X \subset \mathbb{R}^{3}$ as the work space. Let $X_{surf} \subset X$ denote the areas close to the ground, $X_{obs} \subset X$ denote occupied area of space and $X_{free}=X \backslash X_{obs}$. Let $X_{trav} \subset X_{free}\cap X_{surf} $ denote the subspace that is traversable for ground robots.

 
The formal definition of the problem is as follows: Given the initial and target state $x_{start}, x_{goal} \in X_{trav}$, search a feasible control strategy $\pi^{*}$. Input $x_{start}$ to $\pi^{*}$ and it controls the robot to move from $x_{start}$ to $x_{goal}$. $\pi^{*}$ should satisfy: 1) all kinematic and dynamic constraints; 2) avoiding collision with obstacles along the way; 3) minimizing the time needed to move; 4) minimizing the risk of the robot being unable to maintain a stable posture.

\subsection{System Framework}
\ 
Fig.\ref{fig_framework_1} shows the structure of the PUTN algorithm. The SLAM module can be implemented by A-LOAM \cite{zhang2014loam}, Lego-loam \cite{shan2018lego}, etc. The Global Planning module and the Gaussian Process module generate the sparse path and the dense path, respectively. In addition to following the global path, considering that the loss of accuracy and the hysteresis exist in the two modules in real-time navigation, point cloud updated by lidar in real time will be combined in the Local Planning module to improve the responding ability to avoid the obstacles.

\section{Implementation}
\label{sec:implementation}

\subsection{Global Planning}
\label{subsec:global planning}

\subsubsection{Plane Fitting RRT* algorithm}
\ 
\newline
\indent 
PF-RRT* is proposed to solve the problem of global path generation in uneven terrain. Its framework is based on informed-RRT* \cite{gammell2014informed}, and is presented in Alg.\ref{alg:PF-RRT*}. The following are the relevant definitions:
    \begin{itemize}[leftmargin=*]
        \setlength{\itemsep}{0pt}
        \setlength{\parsep}{0pt}
        \setlength{\parskip}{0pt}
        \item $\textbf{x} \in \mathbb{R}^3 $ represents a space point.
        \item $\ob{x}\in \mathbb{R}^2$ represents a plane point.
        \item $\tb{x} \in \mathbb{R}^3$ represents a space point on the surface of the terrain.
        \item $\mathbf{\hat{x}} \in \mathbb{R}^3$ represents the position of the robot center on the terrain.
        \item $S$ represents a path that is a sequence of states.
    \end{itemize}
    
\begin{figure}[htp]
    \centering
    \includegraphics[width=8.5cm]{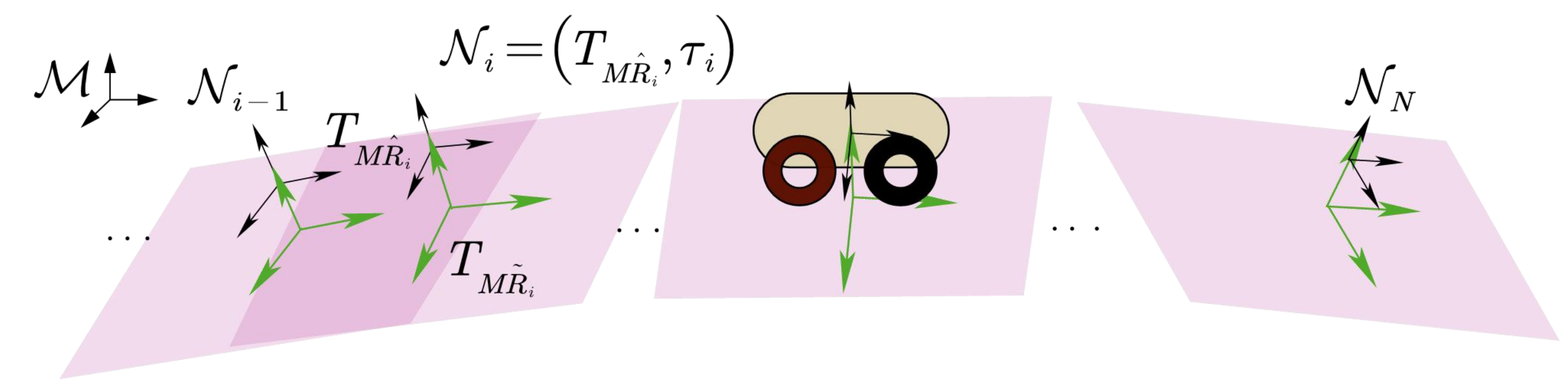}
    \caption{Representation of trajectory for motion planning in complex terrain. The planner trajectory is defined as a series of nodes extended by the sampling tree. Each node $\mathcal{N} _i$ consists of a 6D pose $ T_{M\tilde{R}_i}$ and a traversability $\tau _i$. Based on the set of selected terrain surface points $\varOmega _i$, the SVD method is used to fit a plane $P_i$. Keeping the distance between the nodes satisfies kinematic vehicle constraints.}
    \label{fig:6}
    \vspace{-0.2cm}
\end{figure}

As shown in Fig.\ref{fig:6}, different from the traditional RRT, each element in the tree is represented as a node $\mathcal{N}_i=\left( T_{M\tilde{R}_i}, \tau_i \right)$ instead of a position $\textbf{x}_i$. $T_{M\tilde{R}_i}$ represents a local plane at the center of the robot footprint and $\tau_{i} \in [0,1]$ represents the traversability. A higher value of $\tau_{i}$ means harder to traverse. Here 0 represents absolutely traversable and 1 represents absolutely unable to traverse. The detailed calculation will be introduced in Section\ref{subsec:Terrain Assessment}.

Based on the definition above, given two nodes $\mathcal{N}_{1}$, $\mathcal{N}_{2}$, let $l_{1,2}$ denote the Euclidean distance between them, and the cost function of the line connecting $\mathcal{N}_{1}$ and $\mathcal{N}_{2}$ is defined as follows:  

\begin{equation}
\begin{aligned}
f(\mathcal{N}_{1},\mathcal{N}_{2})=\left(1+ \omega \left(\frac{1}{1-\tau_{1}} + \frac{1}{1-\tau_{2} } -2 \right)\right)\cdot l_{1,2}\\
\end{aligned}
\end{equation}

Where $\omega$ is a penalty scale factor. In this way, the algorithm will aim to not only shorten the path but also avoid those paths that are hard to traverse. However, the traversability in this area is assumed to be uniform, which is a rough approach. Further processing is needed to evaluate the traversability of the path more accurately, and it will be described in Section\ref{subsec:gpr}. 




Some new subfunctions presented in Alg.\ref{alg:PF-RRT*} are described as follows while subfunctions common to the informed-RRT* algorithm can be found in \cite{lavalle2001randomized}\cite{karaman2011sampling}\cite{gammell2014informed}:
    \begin{itemize}[leftmargin=*]
        \setlength{\itemsep}{0pt}
        \setlength{\parsep}{0pt}
        \setlength{\parskip}{0pt}
        
        \item $\textbf{Pos}(\mathcal{N})$: Given a node $\mathcal{N}$, the point stored in it is returned.
        \item $\textbf{ProjectToPlane}(\textbf{x})$: Given $\textbf{x}=[x,y,z]^{T}$, it returns $\ob{x}=[x,y]^{T}$.
        \item $\textbf{ProjectToSurface}(\ob{x})$: Given $\ob{x}=[x,y]^{T}$. Let $res$ be the resolution of the grid map, and $z_{l}$ be the lower limit of the height of the grid map. It returns $\tb{x}=[x,y,z]^{T}$ where $z$ satisfies
        $$\min\limits_{z} z$$
        \begin{equation}
        	\text{s.t.}\left\{
        	\begin{aligned}
        	&[x,y,z]^{T} \in X_{obs} \\
        	&[x,y,z+res]^{T} \in X_{free} \\
        	&z-z_{l}=k\cdot res,k \in \mathbb{N}
        	\end{aligned}
        	\right.
        \end{equation}

        
        \item $\textbf{FitPlane}(\tb{x})$: Given a point $\tb{x}$ on the surface, it fits a plane centered on $\tb{x}$ by combining the information of the global map. This process will be described in detail in Section\ref{subsec:Terrain Assessment}.
        
    \end{itemize}
    
    \SetKwFor{For}{for}{\string do}{}
    \RestyleAlgo{ruled}
    \begin{algorithm}[h]
        \caption{PF-RRT*($\mathcal{N}_{start},\mathcal{N}_{goal},k$)}
        \label{alg:PF-RRT*}
        \LinesNumbered

        $V \leftarrow \{\mathcal{N}_{start}\} , E \leftarrow \emptyset , \sigma^{*} \leftarrow \emptyset, \Omega_{goal} \leftarrow \emptyset$\;
        $T=(V,E)$\;
        
        \For{\rm \textit{i=} 1 to \textit{k}}
        {
            \uIf{$S^{*}\neq\emptyset$}{$\ob{x}_{rand}\leftarrow\textbf{SampleEllipsoid}()$\;}
            \uElse{$\ob{x}_{rand} \leftarrow \textbf{RandomSample}()$\;}
            $\mathcal{N}_{nearest} \leftarrow \textbf{FindNearest}(T,\overline{x}_{rand})$\;
            $\ob{x}_{nearest} \leftarrow \textbf{ProjectToPlane}(\textbf{Pos}(\mathcal{N}_{nearest}))$\;
            $\ob{x}_{new} \leftarrow \textbf{Steer}(\ob{x}_{nearest},\ob{x}_{rand})$\;
            $\tb{x}_{new} \leftarrow \textbf{ProjectToSurface}(\ob{x}_{new})$\;
            $\mathcal{N}_{new} \leftarrow \textbf{FitPlane}(\tb{x}_{new})$\;
            \uIf{$\mathcal{N}_{new} \neq \emptyset$ and \rm \textbf{Pos}($\mathcal{N}_{new}) \in X_{trav}$}
            {
                $\Omega_{near} \leftarrow \textbf{FindNeighbors}(V,\mathcal{N}_{new} )$\;
                \uIf{$\Omega_{near} \neq \emptyset$}
                {
                    $\mathcal{N}_{parent}\leftarrow \textbf{FindParent}(\Omega_{near},\mathcal{N}_{new})$\;
                    $V \leftarrow V \cup \{\mathcal{N}_{new}\}$\;
                    $E \leftarrow E \cup \{(\mathcal{N}_{parent},\mathcal{N}_{new})\}$\;
                    $T \leftarrow (V,E)$\;
                    $T \leftarrow \textbf{Rewire}(T,\Omega_{near},\mathcal{N}_{new})$\;
                    \uIf{\rm \textbf{InGoalRegion}($\mathcal{N}_{new}$)}
                    {
                        $\Omega_{goal} \leftarrow \Omega_{goal} \cup \{\mathcal{N}_{new}\}$\;
                    }
                    $S^{*} \leftarrow \textbf{GeneratePath}(\Omega_{goal})$\;
                }
            }
        }
        \KwRet $S^{*}$
    \end{algorithm}
    
    Before global planning, the space is downsampled and represented as a spatial grid map as shown in Fig.\ref{fig_framework_1}. This shortens the time to find the path, but also reduces the accuracy of the path. The problem will be partially solved in \ref{subsec:gpr}.
    
    
    As illustrated in Alg.\ref{alg:PF-RRT*}, the PF-RRT* algorithm repetitively adds random samples to search and optimize the solution. Specifically, the PF-RRT* algorithm samples and expands with 2D methods to get a new plane point $\ob{x}_{new}$. Then, $\ob{x}_{new}$ is first projected to a surface point $\tb{x}_{new}$. After that, a new local plane centered on $\tb{x}_{new}$ will be fitted. Based on the analysis of the local plane, a new node $\mathcal{N}_{new}$ is generated, storing a new state $\mathbf{\hat{x}}_{new}$ and its corresponding traversability. Next, the PF-RRT* algorithm will connect $\mathcal{N}_{new}$ to the tree and execute the optimization with 3D methods if the validity of $\mathcal{N}_{new}$ is verified by collision-checking. The PF-RRT* algorithm repeats the operations above until it iterates to the specified maximum iterations, and an optimal solution $S^{*}$ will be returned.
        
    Furthermore, the algorithm uses the previous path as a heuristic which ensures the stability of the posture of the robot. In addition to inheriting the fast search and convergence speed of the informed-RRT* algorithm, the PF-RRT* algorithm has the following advantages:
 
First, many traditional algorithms analyze the entire map before planning. However, as the environment enlarges and updates, the process of analysis will tend to be highly time-consuming, which is lethal in the field of real-time planning. To address this issue, the PF-RRT* algorithm carries out terrain analysis during the expansion of the random sampling tree to avoid useless analysis, which reduces calculation workload and speeds up the response of the algorithm.
 
Second, while many traditional algorithms adopt a 2D grid map or 2.5D elevation map \cite{yuan2019multisensor}\cite{xiao2021robotic}\cite{fankhauser2018probabilistic}, in which the height value is used as an evaluation index for obstacles or costs, the PF-RRT* algorithm directly adopts 3D map to represent the environment. This enables PF-RRT* to work in more complex 3D environments.
 
\subsubsection{Terrain Assessment}
\label{subsec:Terrain Assessment}
\ 
\newline
\indent 
Terrain assessment is an important step in the navigation process of ground mobile robots. Compared with the traditional assessment method of traversability for a single point in the terrain, ground vehicles are more interested in the traversability of the nearby surface area while driving.

For each point on the terrain surface $\tb{x}_i$, points on the point cloud map are selected by a cube box with sides of length $l_s$, which is represented as $\varOmega _i=\{ ( \tb{x}_{i}^{j} ) _{j=1:N_i} \} $. Based on this set, the SVD method is used to fit a plane $P_i$, and get the unit normal vector $n_i \in \mathbb{R}^3 $.



The three-dimensional coordinate system of the fitted plane can be defined by three unit vectors $e^x_i,e^y_i,e^z_i \in \mathbb{R} ^3$: 

\begin{equation}
\begin{aligned}
e^z_i=n_i
\end{aligned}
\end{equation}

\begin{equation}
\begin{aligned}
e^x_i=\frac{q_i- k_i e^z_i}{\left\| q_i- k_i e^z_i \right\| _2} \end{aligned}
\end{equation}

\begin{equation}
\begin{aligned}
e_{i}^{\mathrm{y}}=e_{i}^{z}\times e_{i}^{x}
\end{aligned}
\end{equation}
where $q_i = \tb{x}_{i+1}-\tb{x}_{i}$, towards the next node. And $k_i = q_i^T.e^z_i$ is a one-dimensional vector, which represents the projection of $q_i$ onto  $e_{i}^{z}$. So
$
R_{M\tilde{R}_i} = [e^x_i,e^y_i,e^z_i]\in \mathbb{R} ^{3\times 3}
$
$
t_{M\tilde{R}_i} = \tb{x}_{i}\in \mathbb{R} ^{3}
$
which represent the rotations and shifts of the local coordinate system on the plane, respectively. 

In this way, each node $\mathcal{N} _i$ corresponds to a local plane $P_i$. The traversability of the area near individual landing points is often described in terms of slope, gradient, and step height. However, when the vehicle is driving, we often pay less attention to the road conditions in small areas (pebbles, clods, etc.). Instead, we are more concerned about the information of ground fluctuation and flatness. Based on the above considerations, instead of analyzing points directly \cite{chilian2009stereo}, the traversability of the plane is determined by three criteria: the slope $s$, the flatness $f$, and the sparsity $\lambda$:
\begin{equation}
\begin{aligned}
\tau ={\alpha}_1\frac{{s}}{{s}_{{crit}}}+{\alpha}_2\frac{{f}}{{f}_{{crit}}}+{\alpha}_3\frac{{\lambda}}{{\lambda}_{{crit}}}
\end{aligned}
\end{equation}
where ${\alpha}_1$, ${\alpha}_2$, and ${\alpha}_3$ are weights which sum to 1. $s_{crit}$, ${f}_{crit}$, and ${\lambda}_{{crit}}$, which represent the maximum allowable slope, flatness, and sparsity respectively, are critical values that may cause the robot to be unable to move or roll over here. $\tau$ ranges from 0 to 1, the higher the value of $\tau$, the worse the condition of the ground. When $\tau  =0$, it means that the terrain is suitable for vehicles, and when  $\tau  =1$, it means that the ground is completely unsuitable for vehicles. These three indicators are calculated as follows: 
\begin{figure}[htp]
    \centering
    \includegraphics[width=8.5cm]{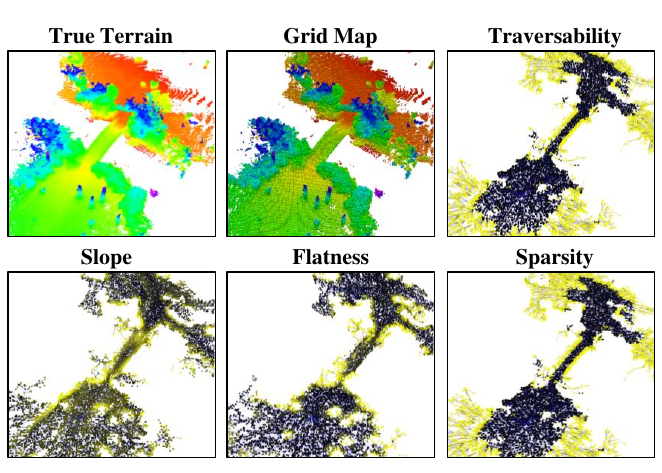}
    \caption{Traversability analysis in the arch bridge area. First downsample the point cloud (upper left) to generate a 3D grid map (upper middle). Then, the RRT tree is expanded and plane fitting and analysis are performed on each node in the 3D grid map. And then the slope, flatness, and sparsity are analyzed respectively (bottom) respectively. These indicators are aggregated to compute the final traversability (top right).}
    \label{pic:traana}
    \vspace{-0.2cm}
\end{figure}

\begin{equation}
\begin{aligned}
s = \mathrm{\kappa}_s \mathrm{arc}\sin  {z}_{e^z}
\end{aligned}
\end{equation}
\begin{equation}
\begin{aligned}
f=\mathrm{\kappa}_f\frac{\sum_{j=1}^N{\left( e^z\cdot x^j \right) ^4}}{N}
\end{aligned}
\end{equation}

\begin{equation}
\begin{aligned}
\lambda =\,\,\left\{ \begin{matrix}
	1&		r>r_{\max}\,\, \\
    \frac{r-r_{\min}}{r_{\max}-r_{\min}}& r \in \left[r_{\min},r_{\max}\right] \land \mathrm{tr}\left( \Sigma ^T\Sigma \right) <t_{\mathrm{trace}}\,\,\\
	0&		\mathrm{otherwise}
\end{matrix} \right. 
\end{aligned}
\end{equation}
where $\mathrm{\kappa}_s$ and $\mathrm{\kappa}_f$ are constant coefficients. ${z}_{e^z}$ represents the projection of the plane normal vector on the Z-axis of the world coordinate system $\mathcal{M}$. $r$ represents the proportion of vacant parts of the plane. $r_{\min}$ and $r_{\max}$ represent the maximum and minimum acceptable vacancy ratio respectively. When the ratio is between the maximum value and the minimum value, whether the vacant points correspond to different situations. When the empty defects are concentrated, it indicates that there may be pits and depressions on the ground. 


As shown in Fig.\ref{pic:traana}, compared with the direct analysis of the landing point, the analysis method of a local plane will take more into account the whole ground situation of the terrain. Since traversability is added to the RRT tree as cost, the RRT tree will automatically stop expanding in difficult-to-pass areas, which makes PF-RRT* more efficient than traditional sampling-based terrain analysis.


In order to ensure a smooth transition between adjacent nodes in the trajectory, it is necessary to ensure that the step length $l_{i,i+1}$ between two nodes satisfies

\begin{equation}
\begin{aligned}
l_{i,i+1}<\frac{l_s}{2}
\end{aligned}
\end{equation}


\subsection{Gaussian Process Regression Prediction}
\label{subsec:gpr}

After global trajectory generation, a trajectory set $S$ and a random sampling tree $T$ are obtained by the PF-RRT* algorithm. Since in each local plane associated with each node, the traversability is assumed to be a constant value, and the sampling step size and planning time limit the density of generated trajectory points. In order to get a denser path, GPR and linear interpolation are used to achieve this. Dense path is defined as 
\begin{equation}
\begin{aligned}
S^I = \left\{ \left( \xi _i^I\,\,\tau_i^I \right) :i\in \gamma_I \right\}
\end{aligned}
\end{equation}
where, $\left\{ \xi_i^I :i\in \gamma_I\right\}$ is obtained by linear interpolation from waypoint set $\left\{ \xi_i:i\in \gamma\right\}$. Here, the method of Gaussian process regression is proposed to predict the traversability $\left\{ \tau_i^I :i\in \gamma_I\right\}$ of interpolation points. The tree node set  $\mathcal{D}=\left\{(\xi_i,\tau_i)_{i=1:N}\right\}$ is used as the training set for Gaussian process regression, and $S^I$ is the test set. The classical radial basis function is adopted as the covariance function
\begin{equation}
\begin{aligned}
k(\xi ,\xi ^{'})=\sigma _{f}^{2}\exp \left[ \frac{-( \xi -\xi ^{'} ) ^2}{2l^2} \right] 
\end{aligned}
\end{equation}



\begin{figure}[htp]
    \centering
    \includegraphics[width=8.5cm]{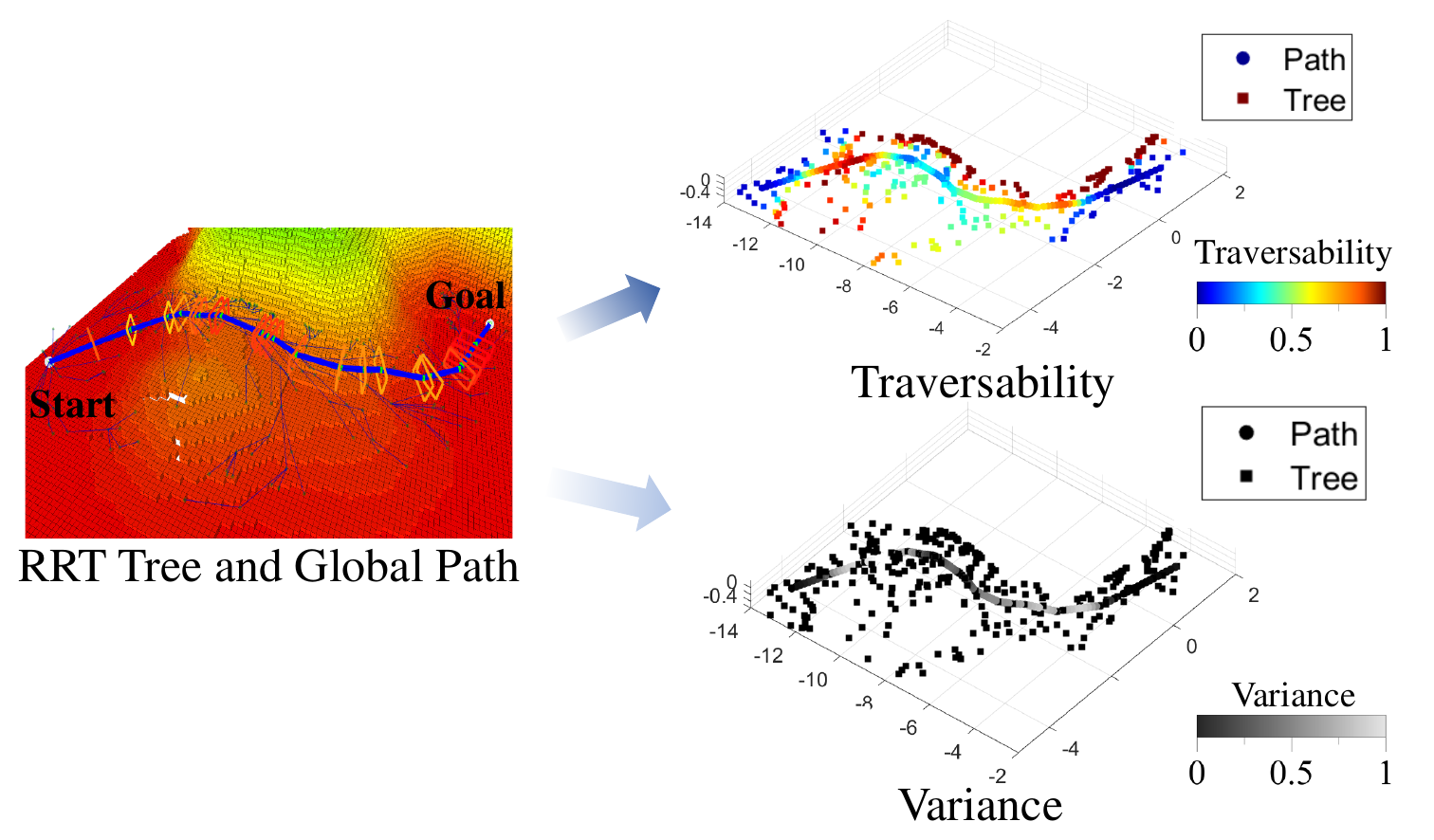}
    \caption{Traversability estimation of interpolated path points using GPR. The global path and the RRT tree generated by the PF-RRT* algorithm on uneven terrain are shown on the left side. The traversability and variance of global path interpolation points estimated by GPR prediction are shown on the right side.}
    \label{fig:estimate}
\end{figure}

Compared with directly using the interpolation of $S$ to calculate the traversability, this method takes the information of the nodes in the RRT tree $T$ and the path $S$ into account. As shown in Fig.\ref{fig:estimate}, the confidence level is greater in areas where nodes are denser, so we can increase the confidence by adjusting the step size of the random exploration.
\begin{figure*}[t]
    \centering
    \includegraphics[width=17cm]{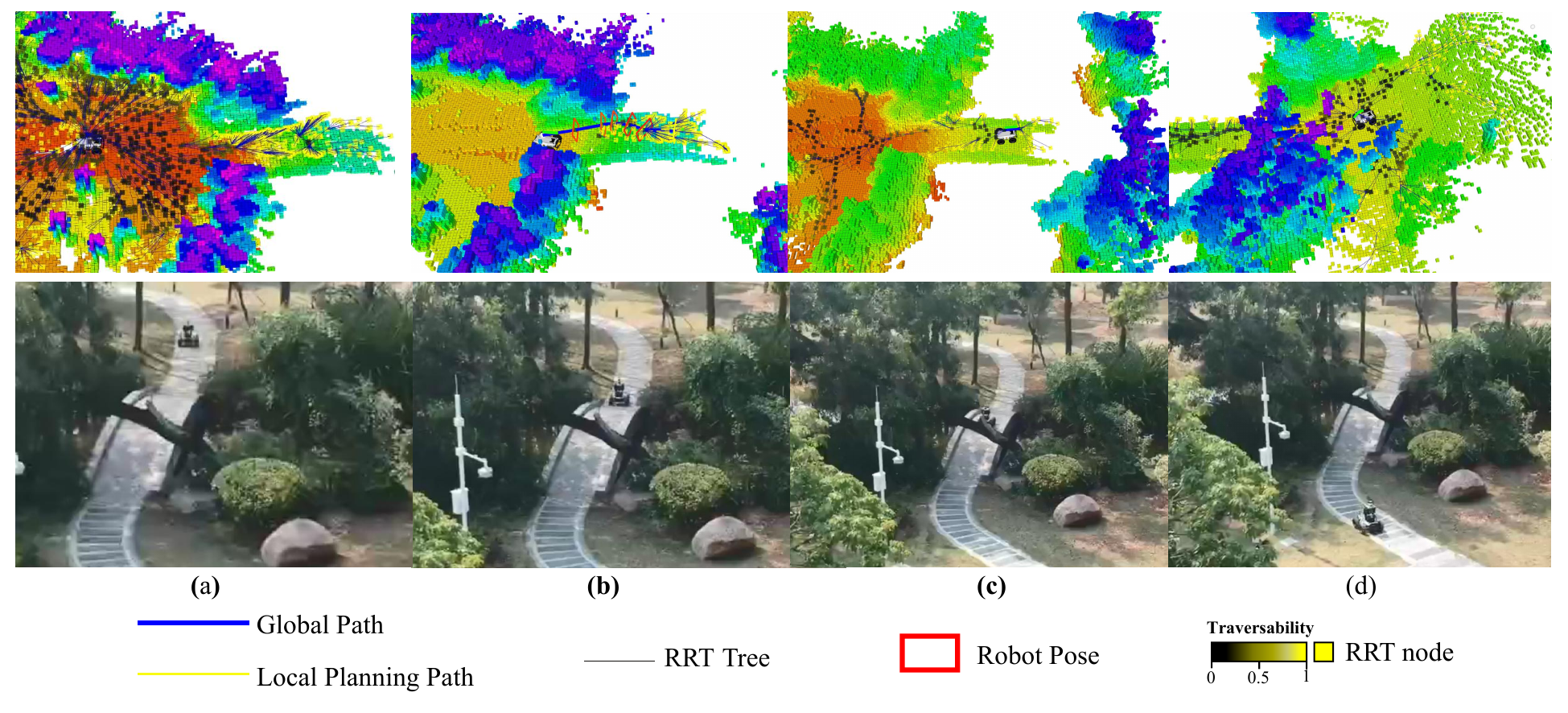}
    \caption{
    PUTN in real scenarios. In the experiment, the starting point and the target point of the robot are located on different sides of the arch bridge. Before the robot moves, the RRT tree continues to expand (a). During the movement (b) (c), the global planner generates temporary goal point close to the final goal point in real time. And the map and the global path are constantly updated. When the robot reaches the end point (d), the RRT starts to expand again until it receives the information of the next target point. The color of each node in the RRT tree indicates the traversability of the current local plane. The RRT tree expands freely on the flat ground and avoids trees or edges of the bridge where are considered to be dangerous. When a global path is found, the local planner guides the vehicle until reaching the goal point.
    }
    \label{pic:arch}
    \vspace{-0.2cm}
\end{figure*}


\subsection{Local Planner}
\label{subsec:localplan}

In local planning, the robot needs to follow the dense global path computed by Global Planning Module and GPR Module. The desired waypoints $S_d = \left\{ \left( p_i,\tau_i,\sigma_i, \right) :i = 0, 1, \ldots, M \right\} $ are selected from the interpolated global path, where $p_i = (x_i^d,y_i^d,z_i^d)$ represents the waypoints, and $\tau_i,\sigma_i$ is the predicted traversability and the uncertainty by GPR.

We formulate the local planner as a NMPC problem. This NMPC leverages a collocation-based trajectory optimization using a simple differential-drive model dynamics to track the given path while considering the traversability and obstacles avoidance. The optimization problem has $N$ nodes and spans over $1$ second. To simplify the problem, we only consider the 2D planning problem in aerial view. The local terrain is described as the normal vector $n_l$ of the local plane with obstacles $\left\{ \mathbf{x}_i^{ob} = \left( x_i^{ob},y_i^{ob} \right) :i = 0, 1, \ldots, N_{ob} \right\} $. We adopt the differential-drive model on local planes as the system model:
\begin{equation}
     \begin{bmatrix}
            x_{k+1} \\
            y_{k+1} \\
            \theta_{k+1} \\
        \end{bmatrix}
        = \begin{bmatrix}
            x_{k} \\
            y_{k} \\
            \theta_{k} \\
        \end{bmatrix} + 
        \begin{bmatrix}
            (\|n_l\times e_x\|)\cos\theta_k & 0 \\
            (\|n_l\times e_y\|)\sin\theta_k & 0 \\
            0&1\\
        \end{bmatrix}
        \mathbf{u}_{k}\Delta{t}
\end{equation}

The optimization formulation has the following form: 

\begin{subequations}
\label{eq:optimization-formulation}
    \begin{align}
        \min_{ \{ \mathbf{x}_k, \mathbf{u}_k \} } \ & \sum_{k=0}^{N-1}{ ( \left\| \mathbf{x}_k - \mathbf{x}_k^d \right\|^2_{Q}
    	+ \lambda  \left\| \mathbf{u_k} \right\|^2_{R} )}
    	  \label{subeq:opti} \\
        \text{s.t.} \ & \mathbf{x}_{k+1}=f(\mathbf{x}_k,\mathbf{u}_k) \label{subeq:dynamics} \\
        & \mathbf{x}_0= \mathbf{x}_{\text{start}} \label{subeq:initial-condition}\\
        & \mathbf{x}_k\in \mathcal{X} ,\mathbf{u}_k\in \mathcal{U} \label{subeq:admissible} \\
        & \left\| \mathbf{x}_k - \mathbf{x}_i^{ob} \right\| \ge d_{\text{safe}} \label{subeq:obconstraint}
    \end{align}
\end{subequations}

Where $\mathbf{x}_k^d$ is two-dimensional vector representing the target points, obtained by $S_d$. $\|\mathbf{x}\|_A := \frac{1}{2}\sqrt{\mathbf{x}^T A \mathbf{x}}$, and the two positive definite matrices $Q$ and $R$ are respectively coefficients measuring terminal costs and control costs. $\lambda = ((1-t_{mean})(1-\sigma_{mean}))^{-2}$ is the coefficient of control cost, $t_{mean}$ and $\sigma_{mean}$ are obtained by averaging $\tau_i$ and $\sigma_i$ in the current local plane, respectively. This coefficient can help the robot slow down when the average traversability is bad or the estimation confidence is low.
\section{EXPERIMENTS}

\begin{figure}[htp]
    \centering
    \includegraphics[width=8.5cm]{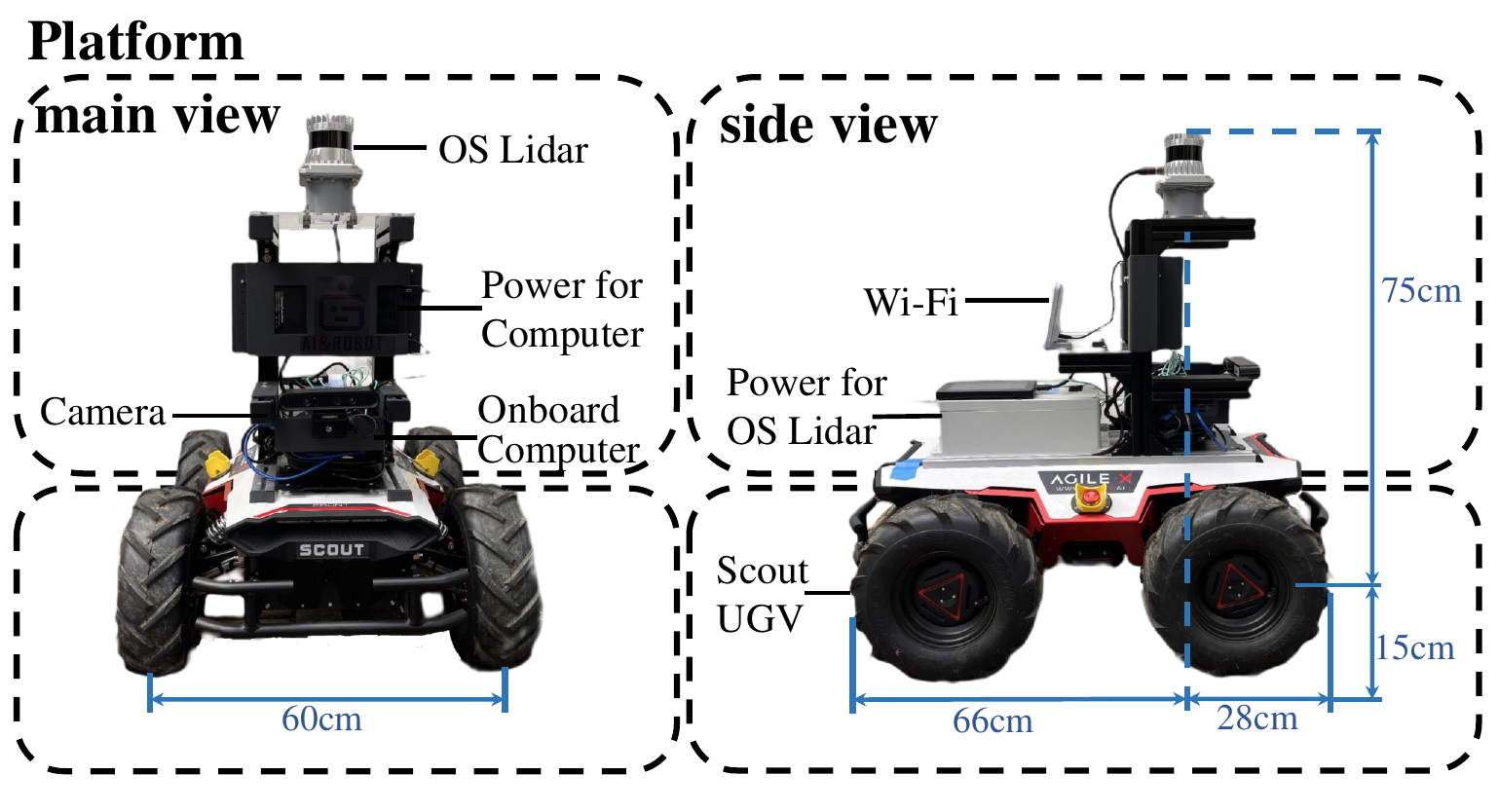}
    \caption{Our PUTN vehicle platform for the experiment, equipped with OS0 lidar and a camera(only for front view in the video). Two batteries are installed to power the lidar and the onboard computer. Wi-Fi is used for communication.}
    \label{fig:UGV}
    \vspace{-0.5cm}
\end{figure}

As shown in Fig.\ref{fig:UGV}, Scout2.0, a four-wheel-drive mobile robot is used for the experiment. The lidar sensor is OS0-128, and the computational hardware is Intel NUC with an i5 2.4GHz CPU and 16GB memory.

\begin{figure}[h]
    \centering
    \includegraphics[width=8.5cm]{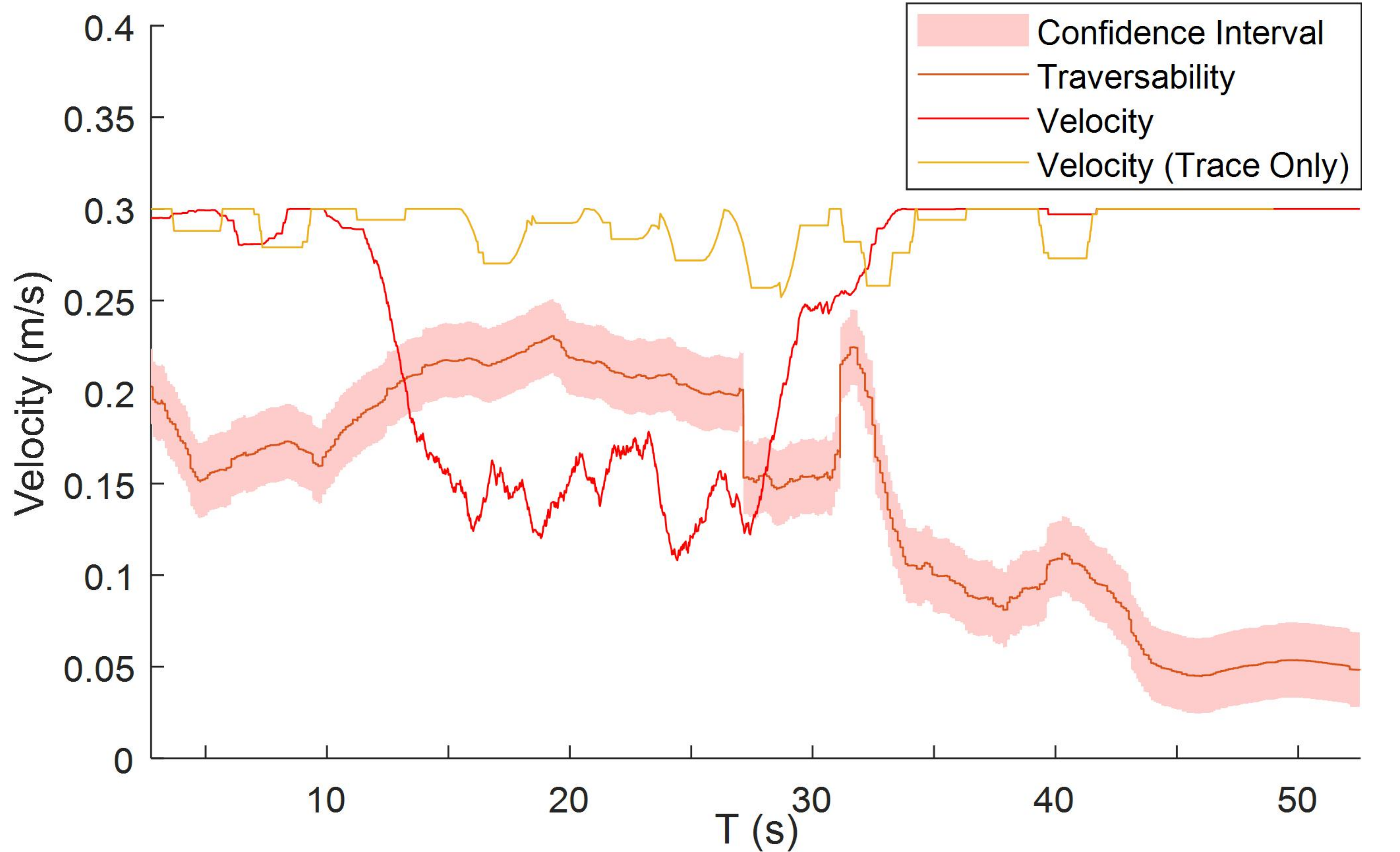}
    \caption{Velocity curve and traversability curve. From the starting point to the target point, the traversability and the corresponding uncertainty of the current vehicle position will change in real time. The linear velocity of the vehicle is recorded with and without traversability in NMPC, respectively.}
    \label{fig:vel}
    \vspace{-0.2cm}
\end{figure}

\begin{figure}[!t]
    \centering
    \includegraphics[width=8.5cm]{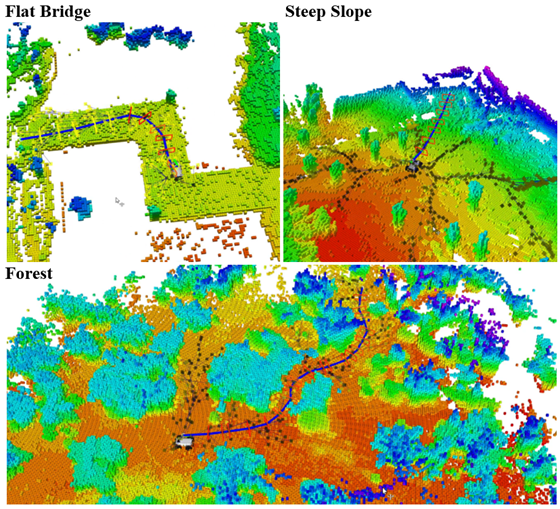}
    \caption{Experiments in flat bridge, steep slope, and forest. The real scenarios corresponding to these experiments can be found in Fig.\ref{fig_exp_sen1}. The video of real-world experiments is available at https://www.youtube.com/watch?v=3ZK-Ut29hLI.}
    \label{fig:exp3}
\end{figure}

Fig.\ref{pic:arch} shows the experiment on arch bridge terrain. CasADi is used to solve the NMPC problem, and a C++ library is used to implement GPR, which takes less time than Python library. As is shown in Fig.\ref{fig:vel}, after traversability and the corresponding uncertainty are added to NMPC, the linear velocity of the vehicle decrease where traversability increase. Through this method, the vehicle travels more smoothly in potentially dangerous areas such as undulating ground and the edge of river. In addition to the arch bridge scenario, we also conduct experiments in the forest, flat bridge, steep slope scenarios, respectively, which are shown in Fig.\ref{fig:exp3}. 

This study also conducted experiments to demonstrate the advantage of the PF-RRT* algorithm in response speed. We compare PF-RRT* with RRT* on normal grid map and grid map analyzed in advance, respectively. The point cloud is analyzed using the same method, which is mentioned in Section\ref{subsec:Terrain Assessment}. The point cloud data are collected from real scenarios, shown in Fig.\ref{fig_exp_sen1}. The same starting point and goal point are selected for each algorithm. Two indicators are used to estimate the performance of each algorithm:

1) Time to find the initial solution. 

2) Time to find the optimal solution. Considering that sample-based algorithms can't guarantee an optimal solution, we compare the time consumption of the two algorithms to reduce the cost to the same value when optimizing the path.


\begin{figure}[!h]
    \centering
    \includegraphics[width=8.5cm]{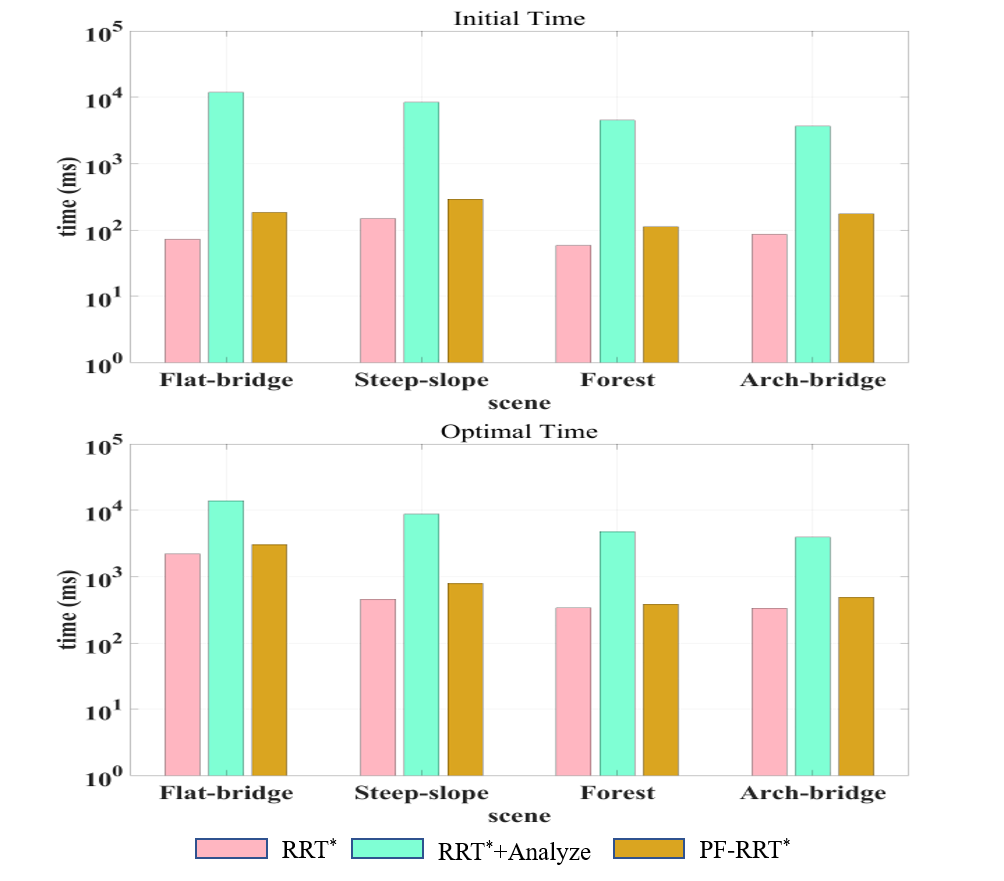}
    \caption{The time comparison among RRT*, RRT*+Analyse, and PF-RRT*. The upper shows the initial time and the lower shows the optimal time. Statistics in each group are obtained from 100 trials.}
    \label{fig:arch}
    \vspace{-0.3cm}
\end{figure}

Fig.\ref{fig:arch} shows the result. RRT* on the map analyzed in advance takes the least amount of time. Since there is no terrain assessment, it represents the minimum time for planning using sampling methods. After adding terrain assessment, RRT* obviously increases a lot of time consuming, while PF-RRT* does not. That's because PF-RRT* only focuses on the terrain where the robot may pass by and completes the task with less terrain analysis. In different scenarios, the accuracy and convergence speed of the algorithm need to be considered comprehensively. PF-RRT* will be faster if the grid map resolution is set to a larger value.


\section{Conclusion}
\label{sec:conclusion}

This study proposes the PUTN algorithm, which is designed for a ground robot to safely and effectively navigate in environments with uneven terrain. A new terrain assessment method is proposed, which integrates different indicators including slope, flatness, and sparsity to evaluate the traversability of terrain. Combined with the informed-RRT* algorithm and this terrain assessment method, a new path planning algorithm, PF-RRT*, is proposed to obtain a sparse global path. By using GPR to regress the RRT tree generated by PF-RRT*, the traversability of the dense path is obtained. PUTN combines PF-RRT*, GPR, NMPC to complete the fast, stable and safe motion planning of the robot. Experiments are conducted in several typical real-world scenarios. The results verify the advantages of the PF-RRT* algorithm and the practicability of PUTN.



{
    \balance
    \bibliographystyle{IEEEtran}
    \bibliography{IEEEabrv, bib/bibliography}

\begin{thebibliography}{10}
\providecommand{\url}[1]{#1}
\csname url@samestyle\endcsname
\providecommand{\newblock}{\relax}
\providecommand{\bibinfo}[2]{#2}
\providecommand{\BIBentrySTDinterwordspacing}{\spaceskip=0pt\relax}
\providecommand{\BIBentryALTinterwordstretchfactor}{4}
\providecommand{\BIBentryALTinterwordspacing}{\spaceskip=\fontdimen2\font plus
\BIBentryALTinterwordstretchfactor\fontdimen3\font minus
  \fontdimen4\font\relax}
\providecommand{\BIBforeignlanguage}[2]{{%
\expandafter\ifx\csname l@#1\endcsname\relax
\typeout{** WARNING: IEEEtran.bst: No hyphenation pattern has been}%
\typeout{** loaded for the language `#1'. Using the pattern for}%
\typeout{** the default language instead.}%
\else
\language=\csname l@#1\endcsname
\fi
#2}}
\providecommand{\BIBdecl}{\relax}
\BIBdecl

\bibitem{putz2018move}
S.~P{\"u}tz, J.~S. Sim{\'o}n, and J.~Hertzberg, ``Move base flex a highly
  flexible navigation framework for mobile robots,'' in \emph{2018 IEEE/RSJ
  International Conference on Intelligent Robots and Systems (IROS)}.\hskip 1em
  plus 0.5em minus 0.4em\relax IEEE, 2018, pp. 3416--3421.

\bibitem{bansal2020combining}
S.~Bansal, V.~Tolani, S.~Gupta, J.~Malik, and C.~Tomlin, ``Combining optimal
  control and learning for visual navigation in novel environments,'' in
  \emph{Conference on Robot Learning}.\hskip 1em plus 0.5em minus 0.4em\relax
  PMLR, 2020, pp. 420--429.

\bibitem{kalogeiton2019real}
V.~S. Kalogeiton, K.~Ioannidis, G.~C. Sirakoulis, and E.~B. Kosmatopoulos,
  ``Real-time active slam and obstacle avoidance for an autonomous robot based
  on stereo vision,'' \emph{Cybernetics and Systems}, vol.~50, no.~3, pp.
  239--260, 2019.

\bibitem{yuan2019multisensor}
W.~Yuan, Z.~Li, and C.-Y. Su, ``Multisensor-based navigation and control of a
  mobile service robot,'' \emph{IEEE Transactions on Systems, Man, and
  Cybernetics: Systems}, vol.~51, no.~4, pp. 2624--2634, 2019.

\bibitem{xiao2021robotic}
A.~Xiao, H.~Luan, Z.~Zhao, Y.~Hong, J.~Zhao, J.~Wang, and M.~Q.-H. Meng,
  ``Robotic autonomous trolley collection with progressive perception and
  nonlinear model predictive control,'' \emph{arXiv preprint arXiv:2110.06648},
  2021.

\bibitem{wang2019safe}
C.~Wang, J.~Wang, C.~Li, D.~Ho, J.~Cheng, T.~Yan, L.~Meng, and M.~Q.-H. Meng,
  ``Safe and robust mobile robot navigation in uneven indoor environments,''
  \emph{Sensors}, vol.~19, no.~13, p. 2993, 2019.

\bibitem{kim2018slam}
P.~Kim, J.~Chen, J.~Kim, and Y.~K. Cho, ``Slam-driven intelligent autonomous
  mobile robot navigation for construction applications,'' in \emph{Workshop of
  the European Group for Intelligent Computing in Engineering}.\hskip 1em plus
  0.5em minus 0.4em\relax Springer, 2018, pp. 254--269.

\bibitem{jeong2021motion}
I.~Jeong, Y.~Jang, J.~Park, and Y.~K. Cho, ``Motion planning of mobile robots
  for autonomous navigation on uneven ground surfaces,'' \emph{Journal of
  Computing in Civil Engineering}, vol.~35, no.~3, p. 04021001, 2021.

\bibitem{xiao2021learning}
X.~Xiao, J.~Biswas, and P.~Stone, ``Learning inverse kinodynamics for accurate
  high-speed off-road navigation on unstructured terrain,'' \emph{IEEE Robotics
  and Automation Letters}, vol.~6, no.~3, pp. 6054--6060, 2021.

\bibitem{fan2021step}
D.~D. Fan, K.~Otsu, Y.~Kubo, A.~Dixit, J.~Burdick, and A.-A. Agha-Mohammadi,
  ``Step: Stochastic traversability evaluation and planning for safe off-road
  navigation,'' \emph{arXiv preprint arXiv:2103.02828}, 2021.

\bibitem{papadakis2012constraint}
P.~Papadakis, M.~Gianni, M.~Pizzoli, and F.~Pirri, ``Constraint-free
  topological mapping and path planning by maxima detection of the kernel
  spatial clearance density,'' in \emph{International Conference on Pattern
  Recognition Application and Methods}, 2012.

\bibitem{fankhauser2018probabilistic}
P.~Fankhauser, M.~Bloesch, and M.~Hutter, ``Probabilistic terrain mapping for
  mobile robots with uncertain localization,'' \emph{IEEE Robotics and
  Automation Letters}, vol.~3, no.~4, pp. 3019--3026, 2018.

\bibitem{campos2021autonomous}
L.~Campos-Mac{\'\i}as, R.~Aldana-L{\'o}pez, R.~de~la Guardia, J.~I.
  Parra-Vilchis, and D.~G{\'o}mez-Guti{\'e}rrez, ``Autonomous navigation of
  mavs in unknown cluttered environments,'' \emph{Journal of Field Robotics},
  vol.~38, no.~2, pp. 307--326, 2021.

\bibitem{zhou2019robust}
B.~Zhou, F.~Gao, L.~Wang, C.~Liu, and S.~Shen, ``Robust and efficient quadrotor
  trajectory generation for fast autonomous flight,'' \emph{IEEE Robotics and
  Automation Letters}, vol.~4, no.~4, pp. 3529--3536, 2019.

\bibitem{shan2018bayesian}
T.~Shan, J.~Wang, B.~Englot, and K.~Doherty, ``Bayesian generalized kernel
  inference for terrain traversability mapping,'' in \emph{Conference on Robot
  Learning}.\hskip 1em plus 0.5em minus 0.4em\relax PMLR, 2018, pp. 829--838.

\bibitem{krusi2017driving}
P.~Kr{\"u}si, P.~Furgale, M.~Bosse, and R.~Siegwart, ``Driving on point clouds:
  Motion planning, trajectory optimization, and terrain assessment in generic
  nonplanar environments,'' \emph{Journal of Field Robotics}, vol.~34, no.~5,
  pp. 940--984, 2017.

\bibitem{lavalle2001randomized}
S.~M. LaValle and J.~J. Kuffner~Jr, ``Randomized kinodynamic planning,''
  \emph{The international journal of robotics research}, vol.~20, no.~5, pp.
  378--400, 2001.

\bibitem{zhang2014loam}
J.~Zhang and S.~Singh, ``Loam: Lidar odometry and mapping in real-time.'' in
  \emph{Robotics: Science and Systems}, vol.~2, no.~9.\hskip 1em plus 0.5em
  minus 0.4em\relax Berkeley, CA, 2014, pp. 1--9.

\bibitem{shan2018lego}
T.~Shan and B.~Englot, ``Lego-loam: Lightweight and ground-optimized lidar
  odometry and mapping on variable terrain,'' in \emph{2018 IEEE/RSJ
  International Conference on Intelligent Robots and Systems (IROS)}.\hskip 1em
  plus 0.5em minus 0.4em\relax IEEE, 2018, pp. 4758--4765.

\bibitem{gammell2014informed}
J.~D. Gammell, S.~S. Srinivasa, and T.~D. Barfoot, ``Informed rrt*: Optimal
  sampling-based path planning focused via direct sampling of an admissible
  ellipsoidal heuristic,'' in \emph{2014 IEEE/RSJ International Conference on
  Intelligent Robots and Systems}.\hskip 1em plus 0.5em minus 0.4em\relax IEEE,
  2014, pp. 2997--3004.

\bibitem{karaman2011sampling}
S.~Karaman and E.~Frazzoli, ``Sampling-based algorithms for optimal motion
  planning,'' \emph{The international journal of robotics research}, vol.~30,
  no.~7, pp. 846--894, 2011.

\bibitem{chilian2009stereo}
A.~Chilian and H.~Hirschm{\"u}ller, ``Stereo camera based navigation of mobile
  robots on rough terrain,'' in \emph{2009 IEEE/RSJ International Conference on
  Intelligent Robots and Systems}.\hskip 1em plus 0.5em minus 0.4em\relax IEEE,
  2009, pp. 4571--4576.

\end{thebibliography}
}

\end{document}